\documentclass[letterpaper, 10 pt, conference]{ieeeconf}
\IEEEoverridecommandlockouts
\usepackage{amsmath} 
\usepackage{amssymb}  
\usepackage{glossaries}
\usepackage{siunitx}
\usepackage{cancel}
\usepackage{placeins}
\usepackage[dvips]{graphicx}
\usepackage{caption}
\usepackage{subcaption}
\usepackage{colortbl}
\usepackage{xcolor}
\usepackage{url}
\usepackage{adjustbox}
\usepackage[showonlyrefs]{mathtools}
\usepackage{pifont}
\usepackage{comment}
\usepackage{booktabs}
\usepackage{mdframed}
\usepackage{adjustbox}
\usepackage{authblk}
\usepackage{color}
\usepackage{bm}
\usepackage{algorithm}
\usepackage{xurl}
\usepackage{cite}
\makeatletter
\let\NAT@parse\undefined
\makeatother
\usepackage{float}
\usepackage{pifont}
\usepackage{hyperref}
\usepackage{algpseudocode}
\usepackage{array}

\author{}

\author{ Pranav Nedunghat$^1$, Jeffrey Mao, Praveen Merumugumala

\thanks{$^1${\tt\footnotesize email: pranavnedunghat@gmail.com}.}
}

\title{\LARGE \bf 
Drone Navigation for Fast, Safe Emergency Response
}

\begin{document}

\makeatletter

\g@addto@macro\@maketitle{
\setcounter{figure}{0}
\centering
 \includegraphics[width=\textwidth]{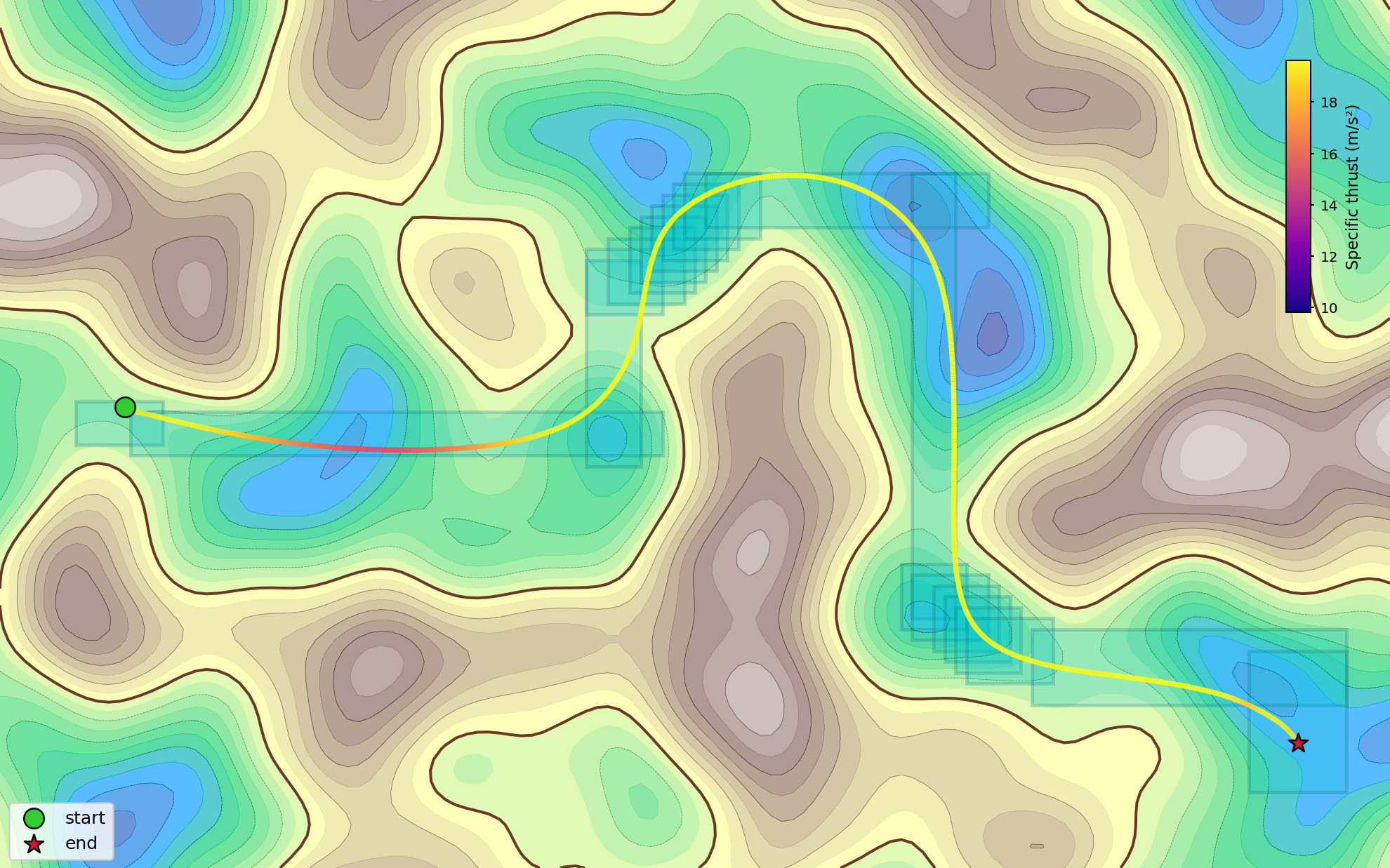}
 \captionof{figure}{Sample Mission. The vehicle must navigate a simulated canyon to reach its goal. Light blue rectangles represent safe flight corridors that the UAV can pass through\label{fig:intro}}
  \vspace{-10pt}
}

\makeatother
\maketitle

\begin{abstract}

 In 2025, there was a total of 358 weather-related disasters worldwide affecting millions of people. For unmanned aerial vehicles (UAVs) to autonomously aid these people, they need to be able to navigate around complex obstacles and reach their destinations to deliver supplies or quickly locate survivors in disaster-stricken areas. Our research addresses this need by providing a method to compute minimum-time trajectories in real time while providing guarantees of obstacle avoidance along the trajectory. Previous approaches treat the problem of computing time-optimal trajectories without considering obstacle avoidance. Our work not only includes obstacle avoidance constraints, but also improves overall mission time for navigation. We define mission time as the sum of planning time and traversal time. In time-sensitive emergencies, rapid trajectory planning is essential to ensure the prompt delivery of critical medical aid. As such, we consider the combined planning and traversal time as the key metric for how quickly the vehicle can reach a survivor. We show a sample mission in Fig. \ref{fig:intro} where the quadrotor navigates through multiple canyon-like obstacles to deliver aid; during most of this trajectory, it is applying the near-maximum thrust.
\end{abstract}

\IEEEpeerreviewmaketitle
\section{Introduction}
Aerial robots have seen success in search and rescue~\cite{Michael_JFR} and the delivery of vital medical supplies. These tasks need to be accomplished as quickly as possible while safely navigating obstacles in complex environments. Our research focuses on generating time-optimal trajectories in real time while also respecting safety constraints and accounting for drag. A key challenge is that time-optimal trajectory planning methods can require multiple seconds [5], [6] of computation each planning instance. Long trajectory-computation times can substantially delay mission completion, even when the resulting flight path is fast. For this reason, practical performance must account for both traversal time and planning time.

Other works have addressed this problem by using a simplified model such as a point-mass model (PMM). For example, recent work \cite{teissing2024real} uses separate acceleration limits for each axis to compute minimum-time trajectories in real time. However, constraining acceleration independently along each axis does not fully utilize the vehicle’s collective thrust capability and can therefore produce suboptimal trajectories. This simplification can produce unnecessarily conservative and therefore slower trajectories.
Our work seeks to bridge the gap between accurate methods with long computation times and fast methods that rely on simplifying approximations. We propose a time-optimal trajectory formulation that accounts for drag, actuator thrust limits, and obstacle avoidance while retaining the computational speed needed for real-time planning. Building on the framework proposed by Mao et al.~\cite{mao2026time}, this work uses Bernstein polynomials to formulate our time-optimal trajectory, model drag dynamics, and formulate the thrust constraint. Using Bernstein polynomial basis functions allows the problem to be solved more quickly, but polynomial representations are often considered poorly suited for time-optimal trajectories. This is because, in a straight line, the time-optimal trajectory exhibits bang-bang behavior, whose discontinuous switching cannot be represented exactly by a finite-order polynomial. Our method addresses this issue by representing the trajectory as a sequence of polynomial splines and optimizng the endpoints over the coefficients. This representation allows our method, as shown in Fig.~\ref{fig:intro} to approximate bang-bang trajectories while incorporating drag and multi-axis collective-thrust constraints.

In this paper, we provide the following key contributions:

\begin{itemize}

\item We provide a time-optimal trajectory-generation method that considers both collective-thrust and obstacle-avoidance constraints using Bernstein polynomials.

\item We propose a method to modify a polynomial trajectory representation to incorporate drag dynamics directly into the optimization. This method is applicable to both Bernstein and regular polynomials.

\item We demonstrate that our approach achieves superior mission time compared with other state-of-the-art-methods \cite{teissing2024real, foehn2021time}  when both planning time and traversal time are considered. 

\end{itemize}

\section{Related Works}\label{sec:related-works}

Existing works on UAV path-planning and computing trajectories can be divided into two main families of approaches. The first approach is to utilize sampling methods such as Probablistic Road Maps~\cite{kavraki1996probabilistic}, Randomly-Exploring Random Trees~\cite{lavalle2001randomized} and its variations ~\cite{webb2013kinodynamic}. These methods typically sample the search space given an initial and final state, which is progressively explored until a solution is reached. While effective for small maps, they quickly become computationally expensive as the search space dimension grows. Moreover, the trajectories computed are not necessarily minimum-time. The second approach is optimization based, such as in ~\cite{mellinger2011minimum} and ~\cite{hargraves1987direct}, where polynomials are utilized to represent continuous time trajectories. This is made possible by the fact that quadrotor dynamics can be formulated as differentially flat systems, thus a trajectory in x, y, z and yaw dimensions can be sampled from these polynomials. However, these approaches neither take obstacle avoidance nor time optimization into account, and can result in the computation of suboptimal or potentially dangerous trajectories.

There have also been many proposed solutions to the problem of computing time-optimal trajectories. Li et al~\cite{lai2006time} proposes a solution that is time-optimal in discretized, but not continuous, time space. Pontryagin's Minimum Principle (PMP) requires that a truly time optimal trajectory should have bang-bang characteristics in velocity, acceleration and/or thrust ~\cite{ritz2011quadrocopter}. Following this, Foehn et al ~\cite{foehn2021time} proposed an approach that has served as a baseline for time-optimal trajectory generation, through the concept of complementary progress constraints (CPC). However, a well-known drawback of this method is the long computation time, in minutes or hours, thus relegating it to be an offline method. A real-time approach was proposed by Teissing et al.~\cite{teissing2024real}, where velocity esimates are initialized and optimized at the waypoint positions. While time-optimal, these approaches neglect the consideration of collision avoidance. 

Multiple methods have been proposed to solve for the minimum time trajectory given a set of obstacles. Penicka et al.~\cite{penicka2022minimum} proposed an approach to generate safe and optimal trajectories by time using a sampling-based strategy to select the best set of velocity states. However, the computational complexity increases quadratically as the number of samples per waypoint increases, hence necessitating the inclusion of a receding horizon of waypoints by Romero et al.~\cite{romero2022time} for real-time replanning. Spedicato et al.~\cite{spedicato2017minimum} proposed a framework for computing minimum-time trajectories through a fixed-horizon spatial formulation, rather than directly optimizing the time variable. However, this reformulation is done using the Frenet-Serret frame, which introduces several complexities, such as nonlinearities and singularites. Furthermore, the use of Euler angles to represent orientation could potentially result in suboptimal solutions. Finally, Mao et al.~\cite{mao2026time}, whose work our approach is built on, shows that Bernstein polynomial based trajectories are competitive with popular baseline methods like MINCO ~\cite{zhou2022swarm} ~\cite{wang2022geometrically} in terms of compute time, while also guaranteeing that safety constraints are respected. However, the previous approach directly constrained velocity and acceleration along each dimension. This constraint formulation was shown to be suboptimal for generating minimum-time trajectories by ~\cite{meyer2023top} and ~\cite{teissing2024real}. Instead, we constrain the thrust norm represented by a 3D sphere, allowing us to utilize the maximum collective thrust of the drone and generate trajectories that are faster than those computed from constraining acceleration and velocity per axis.
\section{Methodology}\label{sec:methodology}
In this section, we introduce our problem statement.
We solve for a trajectory that connects a start state of position, velocity, acceleration and jerk $[\mathbf{x}_0^\top, \mathbf{v}_0^\top, \mathbf{a}_0^\top, \mathbf{j}_0]^\top$ with a goal state $[\mathbf{x}_f^\top, \mathbf{v}_f^\top, \mathbf{a}_f^\top, \mathbf{j}_f]^\top$ given a set of safe flight corridors. In our implementation, we choose the following start and end states: $[\mathbf{x}_0^\top, 0^\top, 0^\top, 0^\top]^\top$ and $[\mathbf{x}_f^\top, 0^\top, 0^\top, 0^\top]^\top$ respectively. 

\subsection{Bernstein Trajectory Problem}
We derive the optimization problem for the single dimensional case, and this can then be extended to an arbitrary number of dimensions. The trajectory is generated using seventh order Bernstein polynomial splines.
First, we define each spline as follows

\begin{equation} \label{eqn:poly_spline}
\begin{split}
B(t) &= 
     \begin{cases}
       \text{$B_{0}\left(t\right)$} &\quad\text{if $t 	\in \left[0,t_0\right]$}\\
        \text{$B_{1}\left(t-t_{0}\right)$}&\quad\text{if $t 	\in \left[t_{0},t_0+t_{1}\right]$}\\
        \hspace{10pt}\vdots\\
        \text{$B_{f-1}\left(t-\sum\limits_{i=1}^{f-2}t_i \right)$} &\quad\text{if $t 	\in \left[\sum\limits_{i=1}^{f-2}t_i ,\sum\limits_{i=1}^{f-1}t_i \right]$}\\
     \end{cases},\\
\end{split}
\end{equation}
where $t_i$ represents the time spent in the safe flight corridor $i$. $f$ is the number of safe flight corridors.  
Each spline $B_i(t)$ is defined as the following function
\begin{equation} \label{eqn:basis_def}
\begin{split}
B_i(t) &= \sum_{k=0}^7 c_{k,i}  b_{k,7}\left(\frac{t}{t_i}\right),\\
\end{split}
\end{equation}
where $c_{k,i}$  is the coefficient of the polynomial, $k$ represents the coefficient number and $i$ is the spline number. We define $b_{k,7}(t)$ as the basis function which is found in \cite{bebot} for order $7$  and 1-D polynomial. We choose polynomials of order 7 since it guarantees that our generated trajectory will be smooth in jerk.

Following ~\cite{mao2026time}, we structure our optimization as solving for a set of waypoints $\mathbf{w} = [\mathbf{w}_0, \mathbf{w}_1,..., \mathbf{w}_{f}]^\top$ and traversal times $\mathbf{t} = [t_0, t_1,...t_{f-1}]^\top$ . 
We define a waypoint as a $\mathbf{w}_i \in \mathbb{R}^{4\times 1}$ with elements $\mathbf{w}_i = \begin{bmatrix}
    B_i(0) & B_i'(0) & B_i''(0) & B_i'''(0)
\end{bmatrix}^\top$ where $B' = \frac{dB(t)}{dt}$ is the velocity, $B''= \frac{d^2B(t)}{dt^2}$ is acceleration, and $B'''= \frac{d^3B(t)}{dt^3}$ is jerk. In our formulation, we have $f$ splines and $f+1$ waypoints describing these splines.
Each spline of the trajectory is completely characterized by the start and end waypoint on the curve. This relationship between coefficients and waypoints is defined as the following for the lower set
\begin{equation}\label{eqn:way_coeff_convert}
\begin{bmatrix}
c_{0,i}\\
c_{1,i} \\
c_{2,i}\\
c_{3,i} \\
\end{bmatrix}
 = \begin{bmatrix}
        1 & 0 & 0 & 0 \\
1 & \frac{t_i}{7} & 0 &0 \\
1 & \frac{2t_i}{7} & \frac{t_i^2}{42}  & 0  \\
1 & \frac{3t_i}{7} & \frac{t_i^2}{14}  & \frac{t_i^3}{210}  \\
    \end{bmatrix} 
\mathbf{w}_{i}.
\end{equation}

We define the upper set of coefficients as

\begin{equation}\label{eqn:way_coeff_1_convert}
    \begin{bmatrix}
c_{4,i}\\
c_{5,i} \\
c_{6,i}\\
c_{7,i} \\
\end{bmatrix} = \begin{bmatrix}
        & 1& -\frac{3t_i}{7} & \frac{t_i^2}{14} & -\frac{t_i^3}{210} \\
 & 1 & -\frac{2t_i}{7} & \frac{t_i^2}{42} & 0  \\
 & 1 & -\frac{t_i}{7} & 0 & 0 \\
 & 1 & 0 & 0 & 0\\
    \end{bmatrix}\mathbf{w}_{i+1}.
\end{equation}

This derivation is extended from the appendix in \cite{mao2026time}.

\subsection{Formulating simplifed Bernstein Time-Optimized Trajectory Planning}
Building on this structure, we formulate the trajectory generation problem with the following optimization 
\begin{equation}\label{eqn:traj_op_problem}
\begin{aligned}
\min_{\mathbf{w}_{[0,...,f]}, \mathbf{t}_{[0,...,f-1]}} \quad &  ~\sum^{f-1}_{i=0}t_i,\\
\textrm{s.t.} \quad \mathbf{w}_0&=\begin{bmatrix}
    \mathbf{x}_0^\top & 0^\top & 0^\top & 0^\top
\end{bmatrix}^\top,\\
\mathbf{w}_{f}&=\begin{bmatrix}
    \mathbf{x}_f^\top & 0^\top & 0^\top & 0^\top
\end{bmatrix}^\top,\\
 \mathbf{y} \leq &\mathbf{G}(\mathbf{t}) \mathbf{w} \leq \mathbf{z},~ 0 < \mathbf{t}    \\
\end{aligned}
\end{equation}

 The matrices  $ \mathbf{y},\mathbf{G}, \mathbf{z}$ are safety constraints based on eq.~(\ref{eqn:ineq_const}). $ \mathbf{w}_0$ is the quadrotor's current position, velocity, acceleration and jerk respectively, $ \mathbf{w}_f$ is the goal position, velocity, acceleration and jerk.  This results in a time-optimized trajectory.
 As mentioned previously by \cite{mao2026time}, formulating the problem using waypoints instead of coefficients improves our optimization in three ways: reducing the number of optimization variables to $5f-4$ instead of $9f$, an easier initialization procedure, and better conditioned gradients.

The safe flight corridors are defined by their position constraints lower $\mathbf{l} = [l_0,l_1,...l_{f-1}]^\top$ and upper $\mathbf{u} = [u_0,u_1,...u_{f-1}]^\top$ for each corridor, in x, y and z dimensions.  With these bounds, we formulate our safety constraints from eq.~(\ref{eqn:traj_op_problem}), $\mathbf{y}$, $\mathbf{G}, \mathbf{z}$, based on the convex hull property of Bernstein polynomials \cite{bebot}. The proof of this property can be found in the appendix in \cite{mao2026time}. The convex hull property states bounds on the coefficients are equivalent to bounds on the full trajectory defined formally as 

\begin{equation} \label{eqn:ineq_const}
  l_i \leq  \mathbf{A}_i\begin{bmatrix}
\mathbf{w}_{i} \\
\mathbf{w}_{i+1} \\
\end{bmatrix} \leq u_i\implies 
 l_i \leq B_i(t)  \leq u_i \hspace{5pt}
\end{equation}
where $\mathbf{A}_i$ is the matrix to map waypoints to coefficient defined in eq.~(\ref{eqn:way_coeff_convert}). 

\subsection{Formulating the Thrust Actuator Constraint and the Drag Dynamics}
The derivative of a Bernstein Polynomial is 

\begin{equation}\label{eqn:bern_deriv_example}  \begin{split}
    B'_n(t) &= \frac{n}{T}
\begin{bmatrix}
b_{0,n-1}(t)  \\
\vdots\\
b_{n-1,n-1}(t) 
\end{bmatrix}^\top\begin{bmatrix}
 -1 & 1 & 0 &   \hdots  \\
 0 & -1 & 1 & \hdots \\
  &  &   \ddots & \ddots   \\
\end{bmatrix}\begin{bmatrix}
c_{0,n} \\
\vdots\\
c_{n,n}
\end{bmatrix}, 
\end{split}
\end{equation}

We derive the thrust polynomial from the position polynomial, and then constrain the Bernstein coefficients of its squared magnitude. Starting, again from the single dimensional case, the position polynomial is defined as:

The required acceleration per axis is defined as 
\begin{equation}\label{eqn:thrust_eqn}
\mathbf{u}_{i}(t) = \mathbf{B}''(t) + [0, 0, g]^\top + \mathbf{D}\mathbf{B}'(t)
\end{equation}
where $\mathbf{D}$ is a diagonal matrix where the diagonal is $[\mathbf{d}_x, \mathbf{d}_y, \mathbf{d}_{z}]$ which are the linear drag coefficient vector and $g$ is the acceleration due to gravity. Bernstein Polynomials have closure in that derivatives, scalar multiplication and adding these objects produce another Bernstein Polynomial with rational coefficients and finite order. For combining two polynomials of different orders, we follow \cite{bebot} method to convert the term $\mathbf{B}''(t)$ to a Bernstein polynomial of a higher order. 

Next, we convert the per axis acceleration into a thrust constraint by taking the norm of the per axis acceleration. This is expressed as $||u_i(t)||^2 < u_{max}^2$ where $u_{max} = \frac{F_{max}}{m}$.
\begin{equation}
u_{i,x}(t)^2 + u_{i, y}(t)^2 + u_{i, z}(t)^2 \leq u_{max}^2
\end{equation}

\begin{figure}[H]
    \centering
    \includegraphics[width=\linewidth]{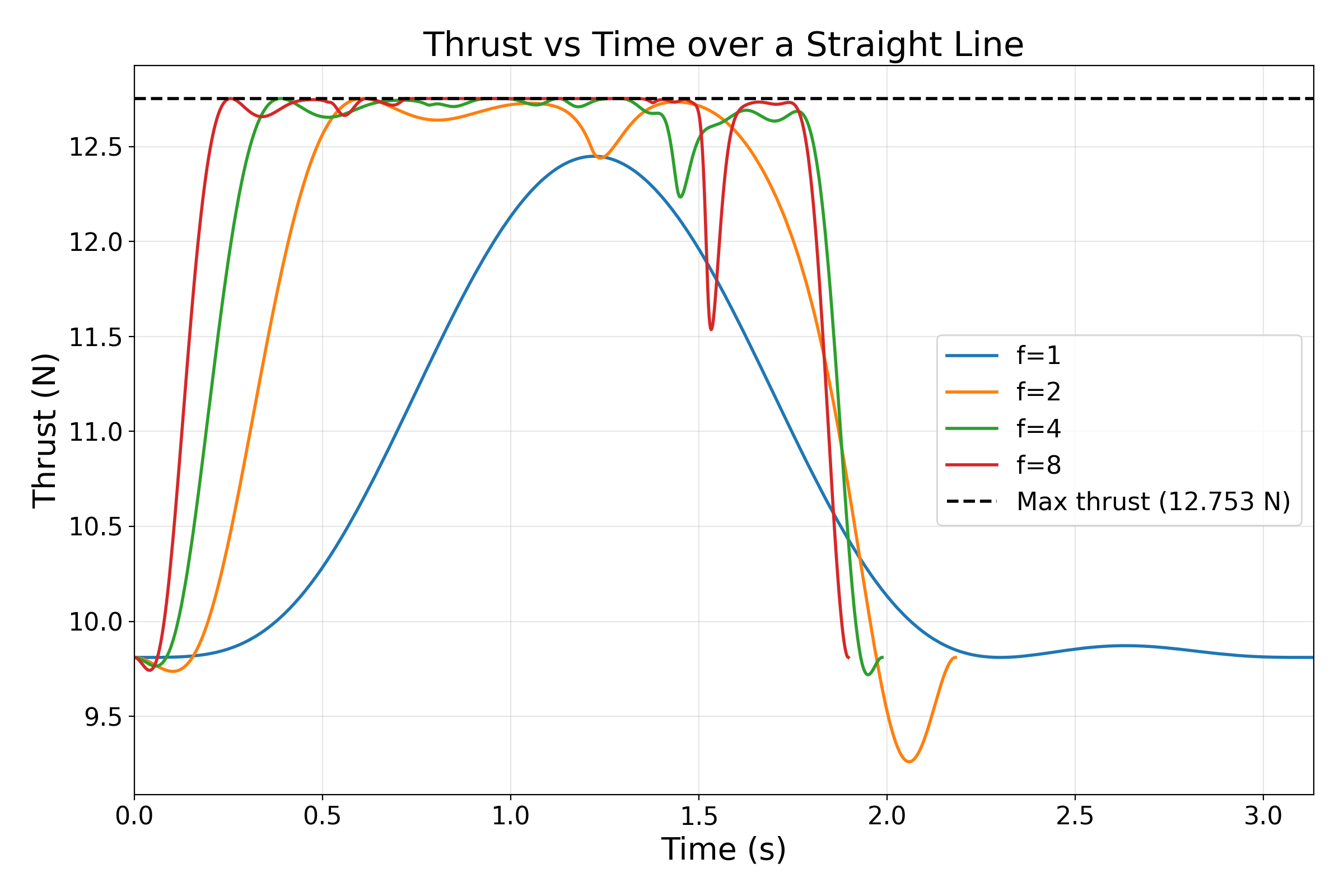}
    \caption{A single spline bernstein polynomial (f = 1) cannot represent max thrust with bang-bang characteristics. In order to better approximate bang-bang characteristics, we need to increase the number of splines representing the thrust curve (f) as can be seen from the figure.}
    \label{fig:hull_certificate}
\end{figure}

 Much like in a normal polynomial, the product of two Bernstein polynomials of order $6$ equation generates a Bernstein Polynomial of degree $12$ equation expressed as follows for the $x$-axis: 
\begin{equation}\label{eqn:degree_12}
\begin{split}
 u_x(t)\cdot u_x(t)
&=\sum_{k=0}^{12}q_{k,x} b_{k,x}\left(\frac{t}{t_i}\right), \\
q_{k, x}&=
\sum_{\substack{l+j=k\\0\leq l,j\leq6}}
\frac{\binom6l\binom6j}{\binom{12}k}
\left(c_{l,x}\cdot\ c_{j,x}\right)\\
\end{split}
\end{equation}

where ${c}_{l,x}$ is index $l$ coefficient for the $x$-axis acceleration. This derivation is valid for the acceleration on all axes $y$ and $z$.
Since this is a Bernstein Polynomial the convex hull property holds, meaning bounds on the coefficients are equivalent to bounds on the polynomial. Thus, we simply need to evaluate the inequalities at the coefficients
Allow $\mathbf{q}_x= [q_{0,x}, q_{1,x}, ... , q_{12,x}]$ which represent the coefficients of our squared axis acceleration. 
This creates the following condition:
\begin{equation}
 \mathbf{q}_x+\mathbf{q}_y+\mathbf{q}_z\leq u_{\max}^2 \rightarrow  ||\mathbf{u}_i||^2 \leq  u_{\max}^2 .
\end{equation}
As such we can simply include $\mathbf{q}_x+\mathbf{q}_y+\mathbf{q}_z\leq u_{\max}^2$
as our thrust constraint in eq.~\ref{eqn:traj_op_problem}. In order to obtain a thrust profile with bang-bang characteristics, we use De Casteljau's algorithm to recursively subdivide the control points, which increases the number of splines representing the thrust curve. The results of this can be seen in Fig ~\ref{fig:hull_certificate}. 

\subsection{Initializing the optimization problem} 
 We then generate the initial trajectory $\mathbf{w}_{\text{init}}$. We assume a rest-to-rest motion profile for each segment of the trajectory. Intermediate waypoints are the intersection of the safe flight corridors or user provided intermediate waypoints. We treat intermediate waypoints as an equality constraint. This trajectory is a straight line trajectory between the start point and the center of the intersections of each safe flight corridors or the next waypoint. To obtain an initial time estimate, we do the following steps:
 \begin{itemize}

\item Double differentiate a seventh order Bernstein polynomial characterised by $w_i$ = $[x_i, 0, 0, 0]$ and $w_{i+1}$ = $[x_{i+1}, 0, 0, 0]$ from ~\ref{eqn:basis_def} to get the acceleration polynomial. Find the peak acceleration of that curve by differentiating and equating the result to 0. This gives us a normalized, dimensionless quantity termed $a_{peak} \approx 7.5132$ for a order 7 polynomial.

\item Similar to our thrust constraint ~\ref{eqn:thrust_eqn} and given specific max thrust $s_{max}$, assume no drag, and solve for maximum acceleration $a_{max}$ from the equation $||a_{max} + g.e_z||^2 = s_{max}^2$

\item $a_{peak} * \frac{L}{T^2}$ = $a_{max}$, where L = $||x_{i+1} - x_i||$ and solve for time T 

\end{itemize}
 Mathematically, we initialize as follows:

 \begin{equation}\label{eqn:init_conditions}
 \begin{split}
        \mathbf{x}_{i} &= \frac{1}{2}(\max     [l_i,l_{i+1}]  +\min [u_i,u_{i+1}]) \cup \mathbf{x}_{provided}, \\
        \mathbf{w}_i &=\begin{bmatrix}
         \mathbf{x}_i^\top & 0 & 0 &0  \\
     \end{bmatrix}^\top \hspace{10pt} 1 \leq i \leq f-1, \\ 
         \mathbf{w}_{\text{init}} &= \begin{bmatrix}
         \mathbf{w}_0^\top & \mathbf{w}_1^\top
        &\hdots &\mathbf{w}_i^\top &  \hdots & \mathbf{w}_{f}^\top\\
     \end{bmatrix}^\top, \\
         \mathbf{t}_{\text{init}} &= \begin{bmatrix}
            \sqrt{\frac{7.532 *||\mathbf{x}_1- \mathbf{x_0}||}{||\mathbf{a}_{max}||}}  &\sqrt{\frac{7.532*||\mathbf{x}_2- \mathbf{x_1}||}{||\mathbf{a}_{max}||}} & \hdots & 
            \end{bmatrix}^\top.
 \end{split}
 \end{equation}

\section{Experimental Results}~\label{sec:experimental_results}
 \vspace{-15pt}
\subsection{System Setup}
~\label{sec:System_Setup}

We run simulation experiments to evaluate our method. In the experiments, we compare our method against those of Foehn et al. \cite{foehn2021time} and Teissing et al.~\cite{teissing2024real}, on example tracks provided in their open-source implementation, as well as one track of our own. We use the MRS UAV simulator \cite{baca2021mrs}, a state-of-the-art simulator, which comes with different drone profiles and controllers. To enable apples-to-apples comparisons, we use the simulator's default implementation of the SE3 Controller. We use the A300 drone model with a mass of 1.21 kg and constrain the planners to a max thrust to weight ratio is 1.6 for all our simulation work. The simulations are performed on a laptop with an Intel Core Ultra 9 275HX (5.4 GHz) Processor. We evaluate the planners on three criteria:

\begin{itemize}

\item Time taken to generate the reference trajectory, referred to as Planning Time, by the planner.

\item Time taken to complete the reference trajectory by the SE3 Controller, referred to as Traversal Time, in the simulation.

\item Mission time, defined as the sum of Planning Time and Traversal Time. 
\end{itemize}

The example tracks have the following names: Hypotrochoid and Eight. Our custom track is called Canyon Track. Since the example tracks do not have any safe flight corridor constraints, we simply utilize one safe flight corridor that covers the whole track to enable comparisons for all waypoints. We also show the results of our method on a custom track called Canyon Track which includes safe flight corridors, and compare them with the above two baselines as well. Since the baselines do not accept safe flight corridor constraints, we simply consider the middle of the overlap between subsequent SFCs as intermediate waypoints instead, similar to how we initialize our method ~\ref{eqn:init_conditions}. Please note the we refer to Foehn et al.'s~\cite{foehn2021time} method as \textbf{CPC}, and Teissing et al's~\cite{teissing2024real} method as \textbf{PMM}, while we refer to our method as \textbf{Bernstein Optimized (BO)}.
\begin{figure}[t!]
    \centering
    \includegraphics[width=\linewidth]{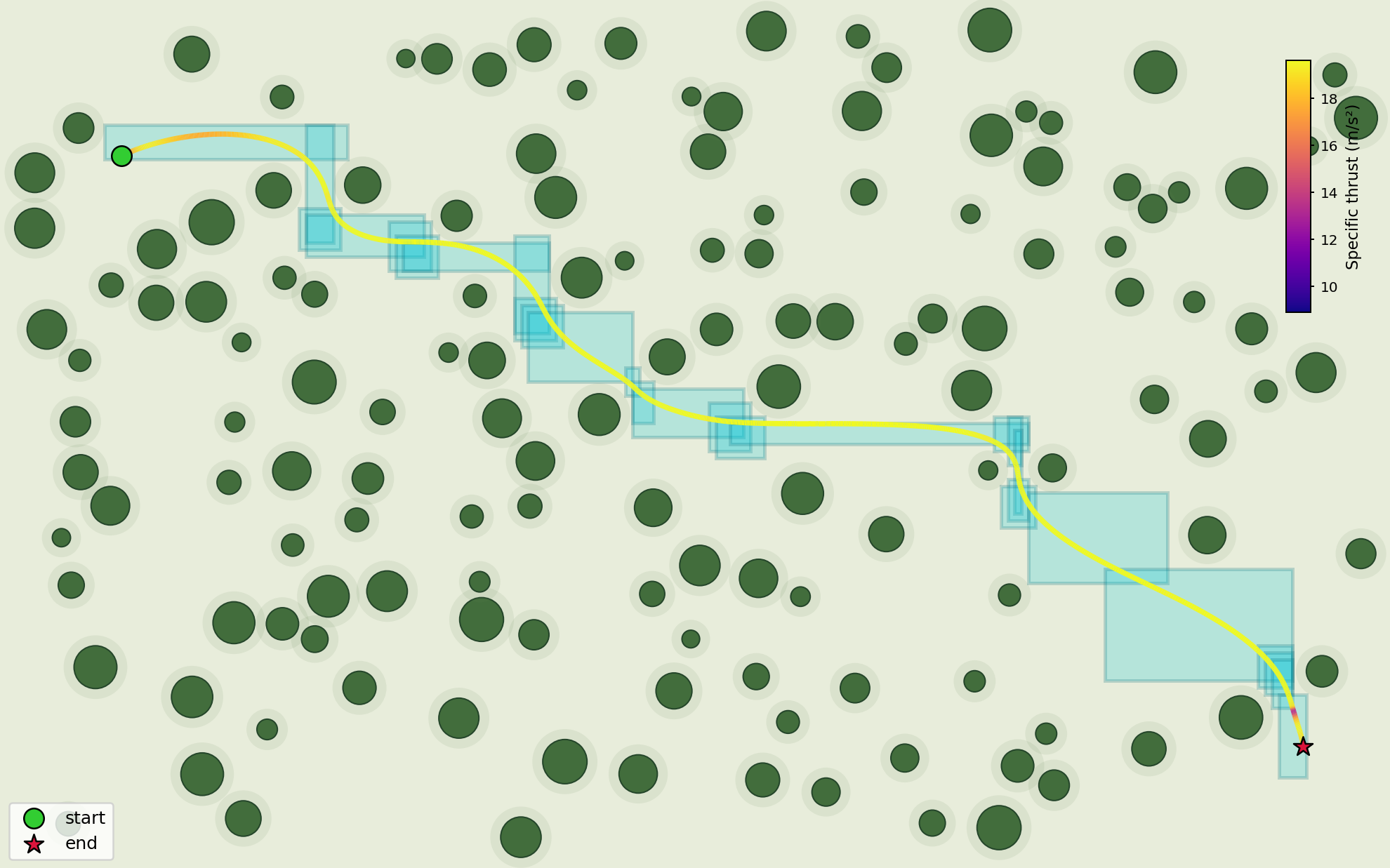}
    \caption{Forest Obstacle Avoidance Trajectory}
    \label{fig:forest}
\end{figure}

\begin{figure}[t!]
    \centering
    \includegraphics[width=\linewidth]{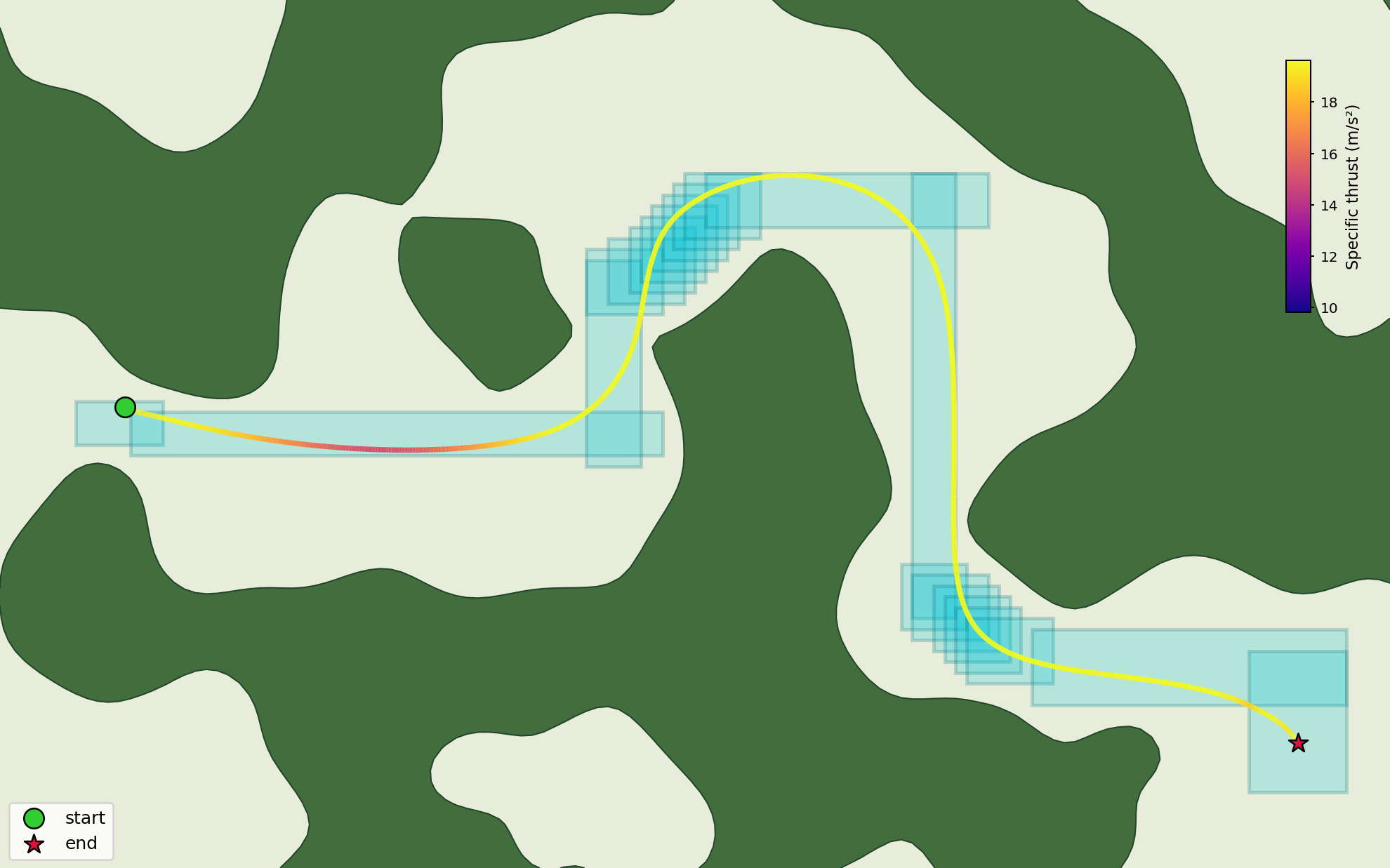}
    \caption{Canyon Obstacle Avoidance Trajectory}
    \label{fig:canyon}
\end{figure}
\begin{figure*}[t!] 
    \centering
    \captionsetup[subfigure]{skip=2pt} 
    
    \begin{subfigure}[b]{\textwidth} 
        \centering
        \includegraphics[width=\textwidth]{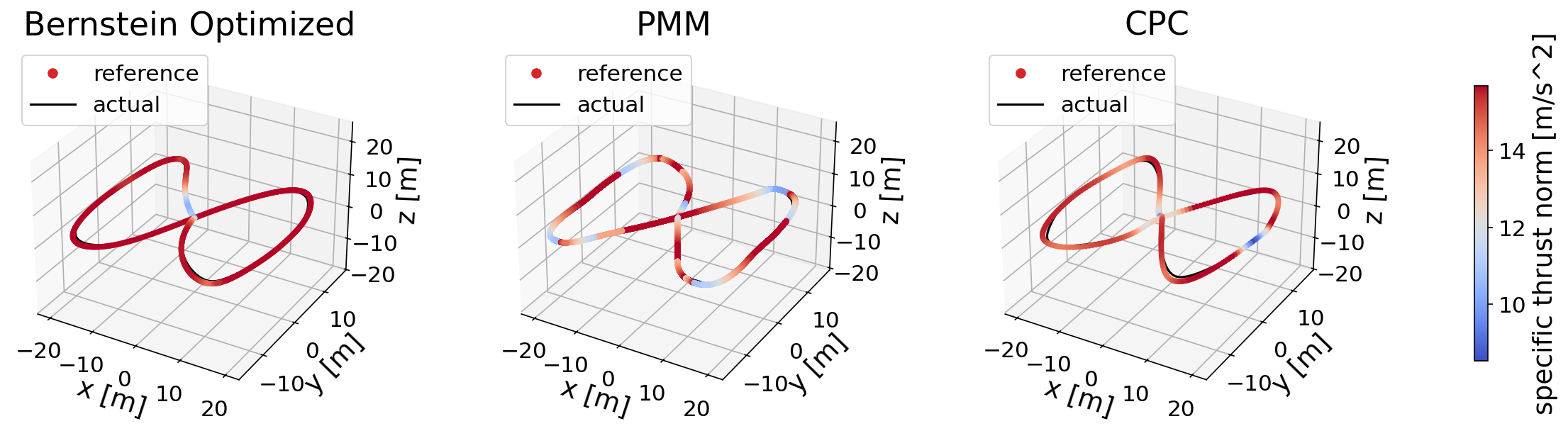}
        \caption{Eight Track}
        \label{fig:sub-a}
    \end{subfigure}
    
    \vspace{0.3cm} 

    \begin{subfigure}[b]{\textwidth}
        \centering
        \includegraphics[width=\textwidth]{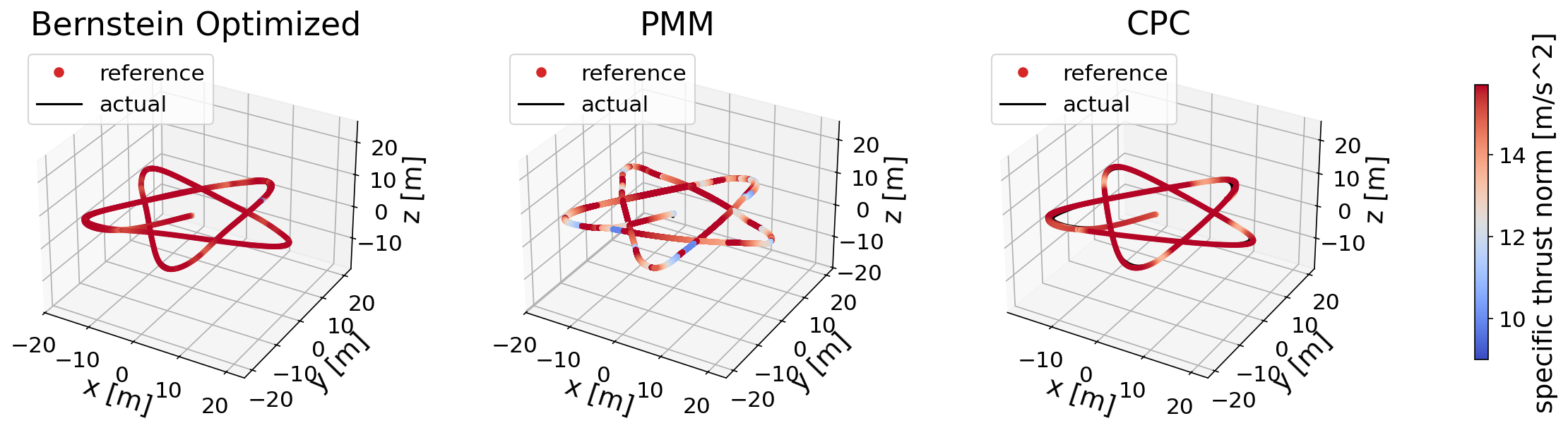}
        \caption{Hypotrochoid Track}
        \label{fig:sub-b} 
    \end{subfigure}
    
    \vspace{0.3cm}

    \begin{subfigure}[b]{\textwidth}
        \centering
        \includegraphics[width=\textwidth]{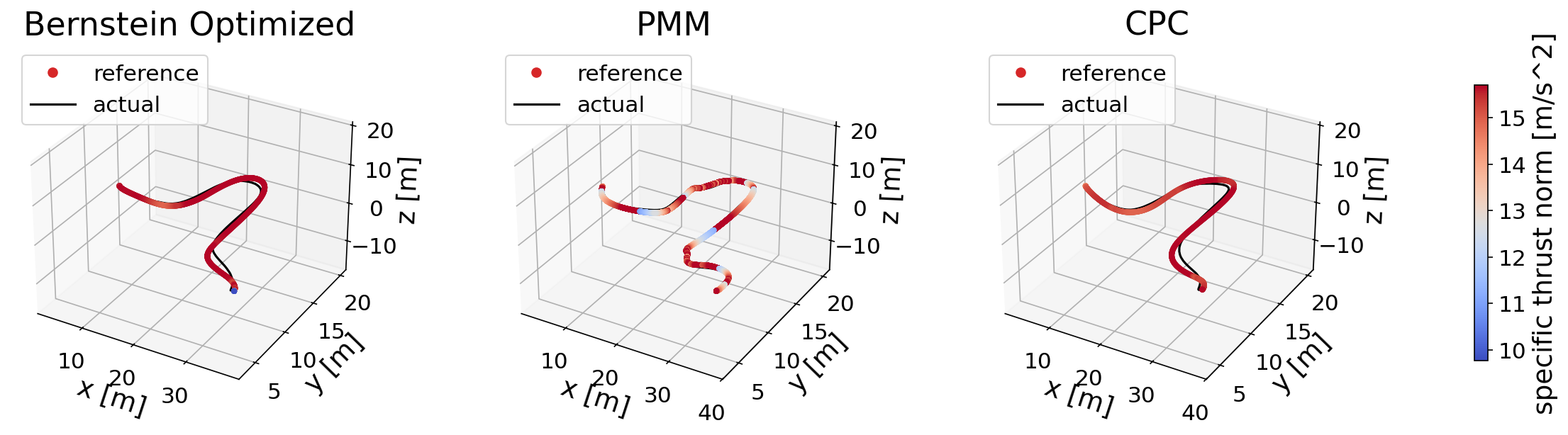}
        \caption{Canyon Track}
        \label{fig:sub-c} 
    \end{subfigure}

    \vspace{0.2cm} 
    \caption{Reference trajectory vs actual tracking during the simulation. The reference trajectory is colored by the specific thrust requested by the planners.}
    \label{fig:tracking-grid}
\end{figure*}

\begin{figure*}[htbp] 
    \centering 
    
    \includegraphics[width=\textwidth]{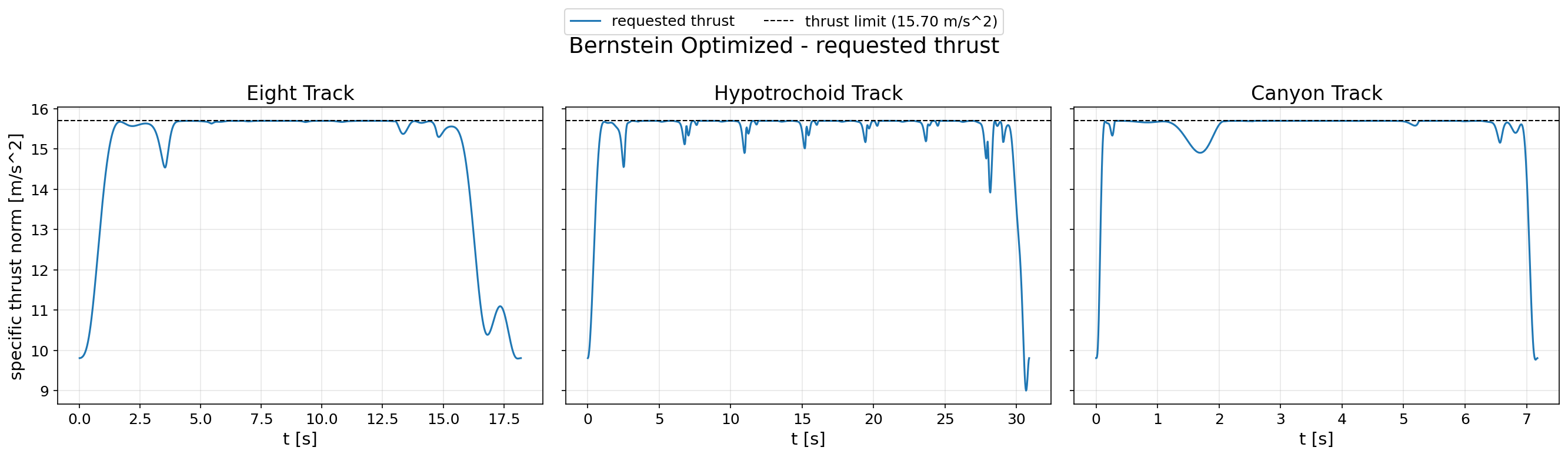} 
    
    \caption{Thrust requested by our method for each trajectory. The dotted line shows max thrust, showcasing that our method approaches bang-bang-like behavior on thrust.} 
    \label{fig:thrust-grid} 
\end{figure*}

\begin{table*}[t]
    \centering\resizebox{\textwidth}{!}{
    \begin{tabular}{|c|c|c|c|c|c|c|c|c|c|}
     \hline
       Method & Planning $(\si{s})$ & Traversal  $(\si{s})$   & Mission  $(\si{s})$  & Planning  $(\si{s})$ & Traversal  $(\si{s})$   & Mission  $(\si{s})$ & Planning  $(\si{s})$ & Traversal  $(\si{s})$  & Mission  $(\si{s})$ \\
       \hline
     &  \multicolumn{3}{|c|}{Eight Track} & \multicolumn{3}{|c|}{Hypotrochoid Track}  & \multicolumn{3}{|c|}{Canyon Track}  \\
        \hline
       BO (Ours)  & $0.125$ & $18.40$ & $\mathbf{18.525}$ & $0.397$ & $31.01$ & $\mathbf{31.4}$ & $1.664$ & $7.46$ & $\mathbf{9.83}$\\
       PMM\cite{teissing2024real} & $0.006$ & $21.85$ & $21.856$  & $0.015$ & $35.05$ & $35.06$ &  $0.012$ & $11.21$ & $11.22$\\
       CPC\cite{foehn2021time} & $94.65$ & $18.26$ & $112.91$& $1451.41$ & $28.81$ & $1480.22$ & $297.221$ & $7.61$ & $304.831$ \\
       \hline
    \end{tabular}}
    \caption{ BO refers to Bernstein Optimized our method. Execution time for the various tracks. Planning refers to planning time which is how long it took to solve the trajectory. Traversal refers to how long it took for the simulator to complete the trajectory. Mission refers to mission time which is the sum of planning time and traversal time.}
    \label{tab:track_table}
\end{table*}

\subsection{Obstacle Avoidance}
In Fig~\ref{fig:forest} and Fig.~\ref{fig:canyon} , we showcase the obstacle avoidance capability of our method. Given a set of safe flight corridors, we guarantee that the trajectory never goes outside their bounds while also maximizing thrust across the entire mission. These safe flight corridors were generated using~\cite{mao2026time} methodology. 
As observed not only are these trajectories safe, they also resemble bang-bang trajectories that seek to maximize the thrust and as a result minimize the traversal time. A further benefit of using safe flight corridors over regular waypoints is the the optimizer has the freedom to move the position traversed anywhere inside the overlapping corridors, rather than being constrained to go through a specific position waypoint. This could potentially reduce travel time even further.

\subsection{Mission Time}
As can be seen from the results in Table \ref{tab:track_table}, our method is able to consistently achieve better total time compared to the two baselines across the tracks. While CPC is able to compute faster trajectories consistently, the compute time to generate the trajectory is too long, and must subsequently be done offline. PMM is able to compute trajectories extremely quickly (within milliseconds), but the resulting trajectory is not as fast as our method or CPC's, thus being slower overall compared to our method.

\subsection{Tracking Performance}
The results of the simulations can be seen in Fig ~\ref{fig:tracking-grid}. It is a well known fact that minimum-time or time-optimal methods generate trajectories that are infeasible for a regular controller, thus requiring specialized controllers such as a Model Predictive Contouring Control (MPCC)~\cite{romero2022model}. However, since the implementation was not available to us, we decided to utilize the SE3 Controller, which has known difficulty tracking aggressive trajectories. We thus had to reduce max thrust-to-weight ratio to 1.6 across all three methods to achieve acceptable tracking performance. This inability to track aggressive motion is the reason behind the greater tracking error for CPC, since CPC's trajectories are usually slightly more aggressive on turns. The position RMSE errors were 0.78 m, 0.96 m, 0.62 m for the Eight, Hypotrochoid and Canyon Tracks respectively. Meanwhile, PMM's less aggressive trajectories resulted in less tracking errors. The position RMSE errors were 0.52 m, 0.53 m, 0.37 m for Eight, Hypotrochoid and Canyon Tracks respectively. Since our trajectories were in between CPC and PMM in terms of aggressiveness, our tracking errors were also in between the two baselines, with position RMSE errors being 0.62 m, 0.64 m, 0.69 m for the Eight, Hypotrochoid and Canyon Tracks respectively.

\section{Conclusion}~\label{sec:conclusion}
In this paper, we introduce a method to compute minimum-time trajectories given a set of safe flight corridors and a max thrust constraint. We demonstrate and compare this formulation against other state-of-the-art time-optimal approaches and show that our method achieves superior mission time while satisfying the safety constraints imposed by the optimization problem. Together, these results establish a trajectory-planning framework that is not only fast enough for real-time operation, but aggressive enough to exploit the vehicle’s performance while maintaining continuous safety guarantees. We show that the thrust profile for our method approaches bang-bang-like characteristics, as well as show that it can traverse safely through designated safe flight corridors. Future work will include modeling the drag as quadratic or cubic in terms of velocity for our method, and validating the framework through live-flight experiments.

\addtolength{\textheight}{0.0cm}   

\bibliographystyle{IEEEtran}
\bibliography{references}

\end{document}